\pdfoutput=1

\documentclass[11pt]{article}

\usepackage[preprint]{coling}

\usepackage{times}
\usepackage{latexsym}

\usepackage{amsmath}
\usepackage{pdflscape}
\usepackage{breqn}

\usepackage{enumitem}

\usepackage[T1]{fontenc}

\usepackage[utf8]{inputenc}

\usepackage{microtype}

\usepackage{inconsolata}

\usepackage{graphicx}

\title{On-Device Emoji Classifier Trained with GPT-based Data Augmentation for a Mobile Keyboard}

\author{Hossam Amer, Joe Osborne, Michael Zaki and Mohamed Afify \\
  Microsoft \\
  \texttt{\{hossamamer, joe.osborne, michael.zaki, mafify\}@microsoft.com}}

\begin{document}
\maketitle
\begin{abstract}


Emojis improve communication quality among smart-phone users that use mobile keyboards to exchange text. To predict emojis for users based on input text, we should consider the on-device low memory and time constraints, ensure that the on-device emoji classifier covers a wide range of emoji classes even though the emoji dataset is typically imbalanced, and adapt the emoji classifier output to user favorites. This paper proposes an on-device emoji classifier based on MobileBert with reasonable memory and latency requirements for SwiftKey. To account for the data imbalance, we utilize the widely used GPT to generate one or more tags for each emoji class. For each emoji and corresponding tags, we merge the original set with GPT-generated sentences and label them with this emoji without human intervention to alleviate the data imbalance. At inference time, we interpolate the emoji output with the user history for emojis for better emoji classifications. Results show that the proposed on-device emoji classifier deployed for SwiftKey increases the accuracy performance of emoji prediction particularly on rare emojis and emoji engagement.

\end{abstract}

\section{Introduction}

Emojis improve communication quality among smart-phone users in different text applications \cite{tomihira2018does, cappallo2018new}. For example, emoji usage in tweets has been steadily increasing over the years leading to a 25\% increase in engagement for tweets that feature at least one emoji \citep{agnew2017emoji, mcshane2021emoji, lopez2017did}. Thus, on-device prediction of emojis describing the user input message's intent is a valuable feature for mobile keyboards \cite{chen2024octopus}. Microsoft SwiftKey is a virtual keyboard app originally developed by TouchType serving many users and languages. This paper targets SwiftKey and proposes an on-device emoji classifier model constrained by decreased memory usage and low-latency while providing quality predictions.

On top of the on-device constraints, emoji classification comes with two unique challenges. First, the possible outputs of an emoji classifier are naturally numerous, context-dependent, and user-dependent. For example, a user inserting the laughing face many times may prefer to see the laughing emoji predicted first in the predictions regardless the message's context. Second, the emoji dataset distribution is naturally imbalanced where emojis such as laughing face appear way more frequently than other rare emojis such as food.

\begin{figure*}[!htbp]
  \centering
  \vspace{-1em}
  \includegraphics[width=1\textwidth]{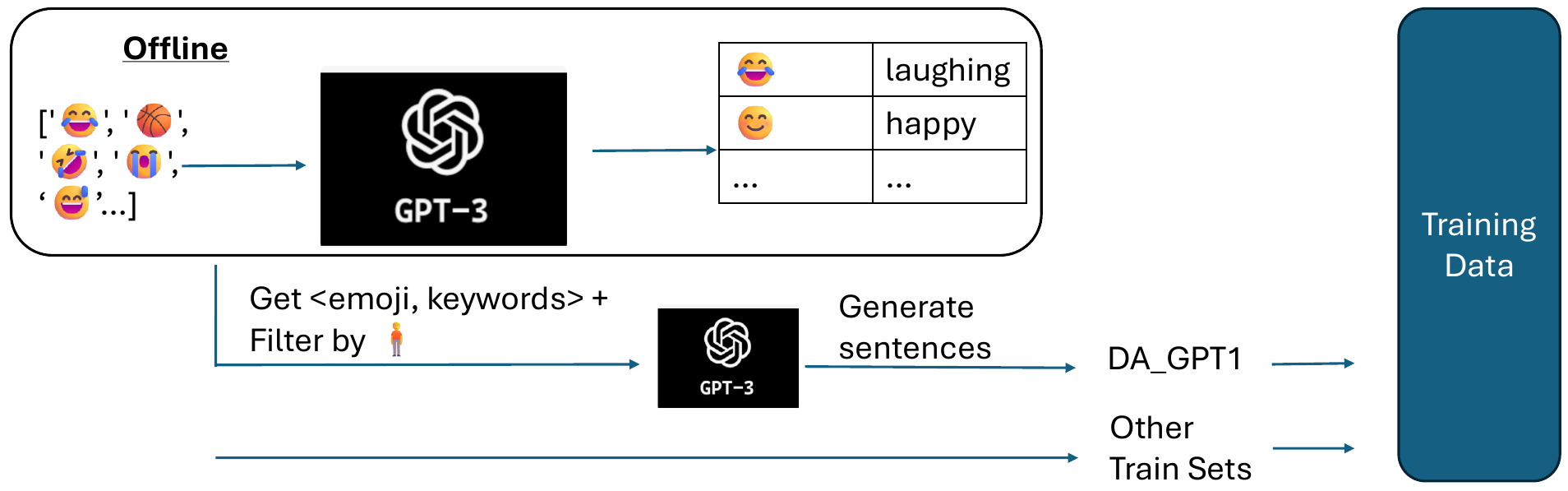}
  \caption{Proposed Data Augmentation Method based on GPT.}
  \label{fig:aug_proposed_method}
\end{figure*}

This paper proposes an on-device emoji classifier trained with a new data augmentation method based on GPT that accounts for imbalanced emoji dataset distribution in numerous emoji classes. First, we provide emoji classification performance to size trade-off comparison between different models that led to selecting a quantized version of the MobileBert transformer as our on-device classifier architecture \cite{sun2020mobilebert}. To tackle the imbalanced emoji distribution, we prompt GPT to create one or more tags for each emoji \cite{brown2020language, lee2021neural, meng2022generating} (see Figure \ref{fig:aug_proposed_method}). Using each of the emoji-tag mapping, we instruct GPT to generate synthetic sentences for this tag and we label the sentence with the associated emoji without requiring any human intervention. The training set is finally a concatenation of this synthetically created keyword set and other human written training sets. Besides, we leverage a method that reranks the proposed emoji classifier output based on the history of emoji user choices at inference time to adapt the output emoji classes to user preferences. 

Experimental results show that the proposed on-device emoji classifier with different number of output classes gains from the proposed augmentation and user history reranking. These offline experimentation ends with the latency required for the on-device emoji classifier and is further verified via live results from real users. Contributions of this paper are summarized below:

\begin{itemize}[noitemsep,nolistsep]
    \item Propose and develop an on-device emoji classifier that respects the on-device memory and latency constraints and runs in SwiftKey. (Section \ref{sec:model_selection}).
    \item Utilize GPT to alleviate the imabalanced nature of the emoji distributions. (Section \ref{sec:augmentation}).
    \item Leverage a method to rerank the emoji classifier outputs with the user history to adapt the output to user choices (Section \ref{sec:user_favorites}).
    \item Deploy the emoji classifier in SwiftKey English with $\approx$10\% CTR increase. (Section \ref{sec:liveResultsSec})
\end{itemize}



Section \ref{sec:relatedWork} is the related work to emoji classification. Section \ref{sec:method} describes the on-device emoji classifier with the data augmentation and reranking methods. Experimental results for this classifier on 3 different datasets and live experiments are in Section \ref{sec:experiments}. Section \ref{sec:conclusion} concludes the paper.




\section{Related Work}
\label{sec:relatedWork}


Different relevant work streams in the literature mitigated both on-device constrains and emoji classification challenges, which could be divided into two classes: output a combination of emojis at once and output a ranked list of emojis based on probability.

The first class focuses on predicting combination of emojis based on the input text \cite{lin2019predict, peng2021seq2emoji, lee2022multiemo}. For example, \cite{peng2021seq2emoji} proposed an encoder-decoder model to predict combinations of 32 emojis all at once. \cite{lee2022multiemo} suggested a different approach by introducing an LSTM architecture to predict combinations of 64 emojis and explored the correlation between emoji and emotion, which showed better performance than \cite{peng2021seq2emoji}. However, they did not consider numerous emojis in their classifications. \cite{peng2023emojilm} showed a successful approach to translate an input text to a combination of emojis while using synthetically generated samples from GPT. However, they did not consider the on-device low-latency and memory constrains for the modelling part. In addition, \cite{kumar2021voicemoji} presented another method to detect emoji boundary in input text and proposed an LSTM based architecture to predict combination of emojis. In Section \ref{sec:model_selection}, we will show that the Bert based architectures perform better than LSTM.

The goal of the second class is to output a ranked list of emojis based on probability from the input text. For instance, \cite{barbieri2017emojis} proposed a Bidirectional Long Short-term Memory Networks (BLSTMs) model to classify among 20 most frequent emojis. Along the same line, \cite{barbieri-etal-2018-interpretable, felbo2017using} introduced an LSTM architecture with label-wise attention targeting the SemEVal 2018 Shared Task \cite{semeval2018task2} and another dataset with at most 200 emojis. The paper for SemEVal task is for predicting the most likely emoji out of 20 emojis to be used along with a tweet, which did not report results for the transformer architectures. \cite{barbieri2020tweeteval, gamal2023federated} introduced emoji classifiers via transformers, but did not consider on-device latency and memory constrains. These constrains were not also covered by \citep{gandhi2022federated} in the Bert-based emoji classifier trained to predict 700 emojis with traditional sampling and cost-sensitive training techniques to account for imabalanced distribution of emoji dataset. Furthermore, \cite{choudhary2018contrastive} proposed Siamese networks and focused on the performance of emoji classification of low-resource languages. \cite{ramaswamy2019federated} successfully implemented an on-device emoji classifier for GBoard's emoji prediction using an LSTM architecture, which is relatively low in computational requirements. Results in this paper were for English with only 100 emoji classes, which are not numerous for industry setting. 


Building on top of the literature, this paper outputs a ranked list of likely emojis and focuses the novelty on: 1) On-device emoji classifier deployed for SwiftKey with large user base, latency/memory being considered; 2) Procedure for emojis class imbalance; 3) Relatively strong performance/analysis on imbalanced and numerous set of emojis tested in an industry setting; 4) Incorporation of dynamic user preferences for personalization.


\section{On Device Emoji Classifier}
\label{sec:method}

\subsection{Model Selection and Quantization}
\label{sec:model_selection}

In this section, we describe a series of experiments undertaken to select a model architecture, ultimately choosing MobileBert with 2 layers as our on-device emoji classifier \cite{sun2020mobilebert}.  




The first experiment trains and tests LSTM, DeepMoji \cite{felbo2017using}, and MobileBert using a mixture of tweets for the 50 most frequent emoji classes \cite{eisner2016emoji2vec, ma2020emoji}. LSTM that is often reported as an emoji classifier in the literature. The MobileBert size is 46MB, LSTM size is 54MB, and DeepMoji model size is 88MB, while MobileBert is getting higher Top-3 Macro F1 (see Table \ref{tab:public50_mobileBert} for F1 scores). With these results, Bert-like architectures \cite{devlin2018bert} and specifically MobileBert is a good candidate for the proposed on-device emoji classifier given its accuracy and size trade-off. 

This observation is also confirmed when we train both an LSTM and Bert architectures on the 20-class SemEval dataset for 30 epochs with the AdamW optimizer and 1e-4 learning rate \cite{gamal2023federated}. SemEval dataset is a widely-used dataset for emoji classification benchmarking and competition that consists of 1.3 million examples of sentences and their corresponding emoji ground-truth. Our experiments along with results shared in \cite{barbieri2018semeval} show that the top-1 F1 score and top-3 F1 score for the Bert and LSTM models on SemEval are 38\%, 30\% and 60\%, 52\%, respectively. We tested GPT3.5 on SemEval in zero-shot and saw 16\% top-1 F1 score. This motivates that Bert-like and particularly MobileBert eligibility for the on-device emoji classifier (See Table 3 in  \cite{semeval2018task2}).

\begin{table}[htbp] \centering
\centering
{\small
\begin{tabular}{|c| c| c|}
\hline 
Model &  Top-1 Macro F1 &  Top-3 Macro F1
\\ \hline
LSTM & 10\% & 24\%
\\
DeepMoji & 12\% & 12\% 
\\
MobileBert & 13.7\% & 27.9\% 
\\
\hline
\end{tabular}
}
\caption{Performance Comparison between MobileBert and other Models on 50 most frequent emojis.}
\label{tab:public50_mobileBert}
\end{table}

To further decrease the size for the MobileBert on-device classifier while assessing its performance on larger classes, we train MobileNet for 5 epochs while taking one layer, 2 layers, and 3 layers from the architecture on another 100-class mixture of tweets (see Table \ref{tab:90C_mobileBert}). MobileBert with 2 layers shows a good accuracy performance at a relatively small runtime memory of 11MB. According to our simulations, this accuracy performance is insignificantly impacted by the int8 quantization while reducing the size further from 11MB to 3MB. With these results, we declare MobileBert with 2 layers as our proposed on-device emoji classifier that we serve in ONNX format for compatibility and ease of deployment \cite{onnxruntime}. To put runtime memory under control, an ONNX library was compiled only given the emoji classifier operators.




\begin{table}[http] \centering
\centering
{\small
\begin{tabular}{|c| c| c|}
\hline 
Model &  Top-3 Macro F1 & Size (MB)
\\ \hline
MobileBert-1Layer  & 7.1\% & 7.2  \\
MobileBert-2Layers & 7.8\% & 11  \\
MobileBert-3Layers & 7.8\% & 14.7 \\
\hline
\end{tabular}
}
\caption{Subset of 100-class MobileBert Transformer Accuracy and Size Trade-off.}
\label{tab:90C_mobileBert}
\end{table}






\subsection{Dataset Construction and Augmentation}
\label{sec:augmentation}

\begin{figure}[htbp]
\centering
\includegraphics[width=0.55\textwidth]{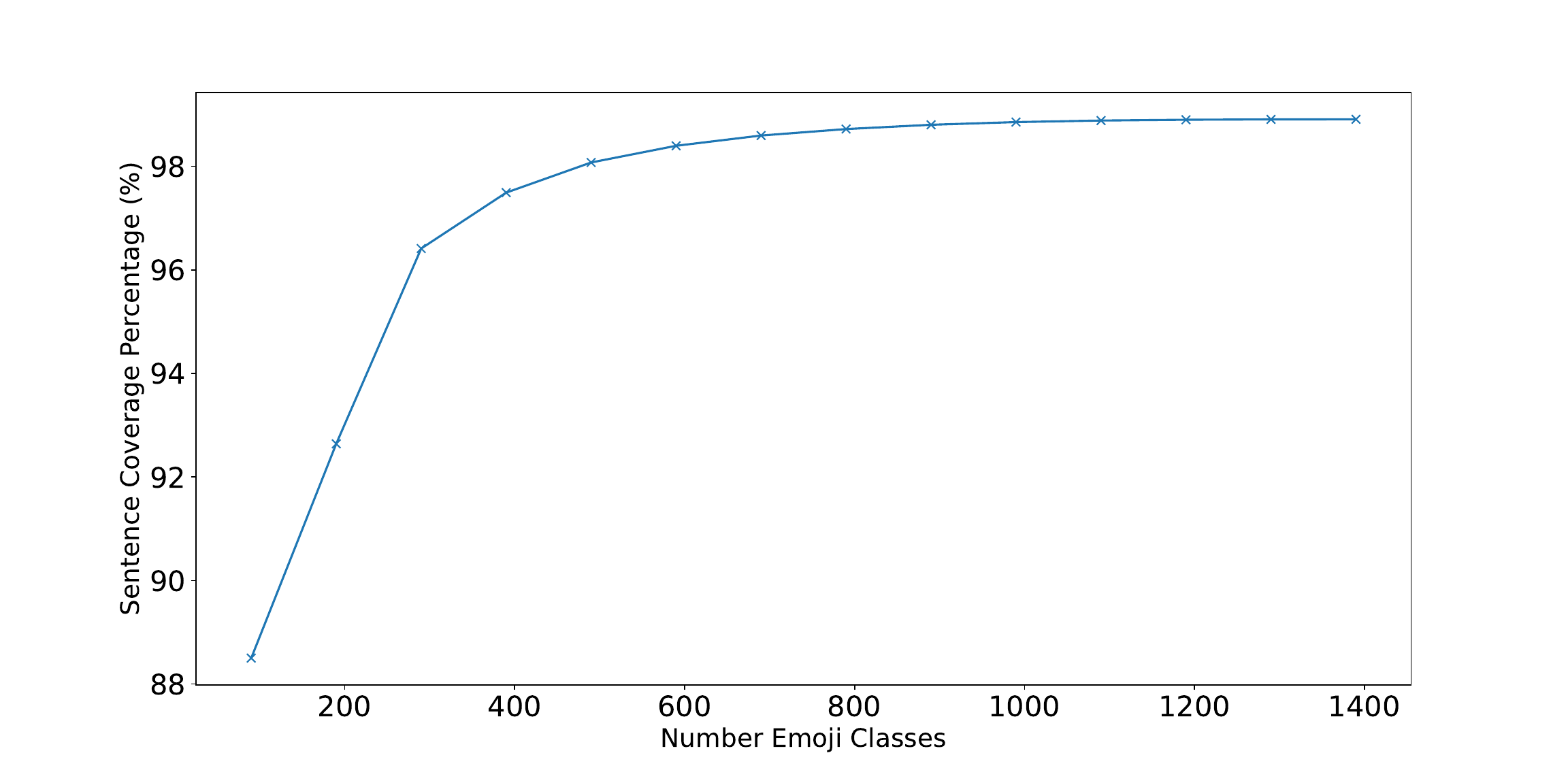}
\caption{Number of Emoji classes versus Percentage of sentences that these emojis are used.}
\label{fig:emoji_coverage}
\end{figure}

As reported in the literature, emojis dataset distribution is imbalanced \cite{ramaswamy2019federated}. This imbalance poses challenges in training an emoji classifier especially when this emoji classifier should cover a large portion of the emoji classes. A user study is conducted to understand this phenomenon. In more detail, we plot in 
Figure \ref{fig:emoji_coverage} the percentage of emoji-containing sentences covered by the Top-K emoji on y-axis and K emojis on the x-axis. These sentences are given by SwiftKey users after taking their consent and removing sensitive information. It's seen that 88.5\% when the emoji classifier classes are 90 (90C), 98.5\% of coverage is achieved with 590C, while it's almost 99\% with 1090C. 


With the shown scarcity of emojis at the tail of the distribution, we 
utilize GPT-3.5 turbo to augment the training data by synthetically generated examples. Figure \ref{fig:aug_proposed_method} shows this method. First, we offline transliterate emojis into their textual meanings using several calls of GPT 3.5 turbo. An emoji may have one or more textual meanings. As soon as we create the emoji to keywords mappings, the authors scan these keywords to reduce any responsible AI or illogical mappings that could come out of the GPT. Then, for each mapping, we instruct GPT to create a sentence with this keyword for which we insert its associated emoji as a label for this sentence. The latter process runs with automatic labeling without human intervention. We believe that the main benefit of this augmentation is not only additional sets particularly for the rare emojis, but also the method's flexibility to add any new uncovered keywords for emojis. Our prompt is presented as follows: 

\textit{Get me a textual description consisting of only one word for <emoji> separated by comma.}

This generated keyword based set is merged with other human written training sets. With this merge, the proposed classifier is trained for two factors simultaneously: emoji transliteration and context understanding for emojis. If a user writes a word in the input that corresponds to an emoji, then it is likely that this user wishes to see this emoji among the output predictions. In addition, the user will want to see emojis describe the overall context of the input sentence.





\subsection{Training Procedure}

We trained the proposed MobileBert with 2 layers in PyTorch via stochastic gradient descent. We utilized GPU training via one NVIDIA A100 GPU with a batch size of 256 for 10 epochs of 3000 steps with learning rate of 2e-4. Weight decay is enabled with decaying factor of 0.01 and maximum sequence length of 50 tokens. The training dataset is a concatanation of synthetic set generated by GPT as explained in Section \ref{sec:augmentation} and  data
logged from SwiftKey users who have opted to periodically share anonymized snippets of text typed in selected apps. All personally identifiable information is stripped from these logs. The logs are filtered further to only include sentences that are labeled as English with high confidence by
a language detection model \cite{botha2017natural, mcmahan2017learning}. The subset of logs used for training contain approximately 12 million snippets containing emoji. The GPT generated set is about 4 million examples. For validation, our user set is about 69,000 samples concatanated with GPT generated sets of 292,000 samples.


\subsection{Emoji Classifier Reranking by User Favorites}
\label{sec:user_favorites}

The output emoji class at inference time should not only depend on the message's context but also the emoji user favorites. For example, a user inserting the laughing face many times may want to see the laughing emoji appear first in the predictions regardless the message's context. For this reason, we rerank the the probability distribution output from the on-device emoji classifier based on the user favorites probability distribution for an input message $m$ as follows:


\begin{multline}
P(emoji|m) = P_{model}(emoji|m) \\\cdot P_{favorites}(emoji)^{\frac{\alpha}{1-\alpha} \cdot \frac{min(s, N_{favorites})}{s}}
\label{eq:user_history_interpolate}
\end{multline}

\noindent where  \( P_{model}(emoji|m) \) is the probability determined by the emoji classifier, and \( P_{favorites}(emoji) \) represents the probability of the emoji being a user favorite. The parameter \(\alpha\) is the reranking coefficient that ranges between 0 and 1.

The probability \( P_{favorites}(emoji) \) is calculated as the number of times an emoji has been inserted divided by the total number of insertion events. \( N_{favorites} \) denotes the number of unique emojis that have been inserted. \( s \), empirically set to 4, is used to lessen the influence of favorites when the number of unique emoji insertions is low.

Equation \ref{eq:user_history_interpolate} enables the favorites probability distribution to re-rank the output of the emoji model, thus adapting to each user's preferences. As \(\alpha\) increases, the impact of the favorites probability decreases while the influence of the emoji model increases, and vice versa. For rare emojis that are not part of the classifier's output, 
\( P_{model}(emoji|m) \) is set to 1e-4, ensuring that user's frequently typed emojis still over time appear in predictions and alleviates the model retraining for new classes. 



\section{Experimental Results}
\label{sec:experiments}


In this section, we present the evaluation results for the proposed on-device emoji classifier with GPT-based data augmentation. We show our results based on three different versions of MobileBert emoji classifier: with 90 classes (90C), 590 classes (590C), 1090 classes (1090C). The emoji classes in the latter models are selected based on popularity in a descending order as indicated by Figure \ref{fig:emoji_coverage}. For evaluation, we utilize three different inhouse testing datasets: D1 with 12426 sentences, D2 with 68804 sentences, and D3 with 1105 sentences. These three datasets are logged from SwiftKey users who have opted to periodically share anonymized snippets of text in selected apps after removal of sensitive information. In particular, D2 is logged from the largest user set (50K). Model quality is evaluated using Hit@1 and Hit@24, defined as the ratio of accurate top-1 and top-24 emoji predictions, respectively, to the total number of examples containing emoji. Hit@24 evaluation metric is selected because SwiftKey displays the top-24 emojis on the predictive panel.



\subsection{Offline Accuracy Comparisons}

Tables \ref{tab:offline_compare_hit1} and \ref{tab:offline_compare_hit24} show the performance for all models under test. We can observe that increasing the number of classes for the on-device emoji classifier lead to increasing the evaluation metrics because there is more coverage for the emoji classes. For example, the 1090C model has a Hit@24 of 10.7\% better than the 90C model. These results go together with the increased coverage shown in Figure \ref{fig:emoji_coverage}. We also use D1 and phi3 zero-shot and saw that phi3 outputs meaningful emojis but may not be necessarily the user’s emoji choice leading to hit1 rates of 8\% plus our models are much smaller.


\begin{table}[http] \centering
\centering
{\small
\begin{tabular}{|c| c| c| c|}
\hline 
Model & D1\_Hit1 (\%) & D2\_Hit1 (\%) & D3\_Hit1 (\%)
\\ \hline
90C & 24.7\% & 12.1\% & 31.2\% \\\hline
590C & 25.4\% & 18.7\%  & 34.3\% \\\hline
1090C & 25.3\% & 20.9\%  & 35.1\% \\\hline
\end{tabular}
}
\caption{Emoji Hit@1 without Augmentation.}
\label{tab:offline_compare_hit1}
\end{table}

\begin{table}[http] \centering
\centering
{\small
\begin{tabular}{|c| c| c| c|}
\hline 
Model & D1\_Hit24 (\%) & D2\_Hit24 (\%) & D3\_Hit24 (\%)
\\ \hline
90C & 70.42\% & 36.7\% & 67.5\% \\\hline
590C & 77.2\% & 54.5\%  & 78.3\% \\\hline
1090C & 77.35\% & 57.0\%  & 78.2\% \\\hline
\end{tabular}
}
\caption{Emoji Hit@24 without Augmentation.}
\label{tab:offline_compare_hit24}
\end{table}

Our evaluation metrics improve further when the proposed data augmentation is added during training for the models under test especially for rare emojis. As indicated in Table \ref{tab:offline_compare_hit_aug}, Hit@1 and Hit@24 reached 21.5\% and 58.2\% on D2 due to the data augmentation on the 1090C model instead of 20.9\% and 57\% without augmentation. 

To further understand these results, we analyze the impact of the data augmentation procedure on the last quarter of rare emojis using D2. The Hit@1 and Hit@24 for D2 on the last quarter of emojis using 1090C without augmentation are 2.5\% and 6.8\%, respectively, while they are 7.1\% and 16.7\% for 1090C with augmentation. Similar results are observed in the 590C that shows the effectiveness of the proposed data augmentation technique especially on rare emojis.




\begin{table}[http] \centering
\centering
{\small
\begin{tabular}{|c| c| c|}
\hline 
Model & D2\_Hit1 (\%) & D2\_Hit24 (\%)
\\ \hline
590C_Aug & 20.5\% & 55.7\%  
\\\hline
1090C_Aug & 21.5\% & 58.2\%  
\\\hline
\end{tabular}
}
\caption{Emoji Hit rates with Augmentation.}
\label{tab:offline_compare_hit_aug}
\end{table}



Turning to user favorites reranking at inference time, Table \ref{tab:offline_alpha_results} shows that increasing $\alpha$ that includes user favorites also increases the Hit@1 and Hit@24 scores for the 1090C. For D2, we see an improvement of 8.9\% in Hit@1 and 11.6\% in Hit@24 at $\alpha$=0.9 relative to 1090C without augmentation shown in Tables \ref{tab:offline_compare_hit1} and \ref{tab:offline_compare_hit24}. This shows that adapting the emoji model responsible for context understanding to each user preferences online lead to better predictive abilities. 

\begin{table}[http] \centering
\centering
{\small
\begin{tabular}{|c| c| c|}
\hline 
$\alpha$ & D2\_Hit1 (\%) & D2\_Hit24 (\%)
\\ \hline
0.1 & 23.5\% & 60.0\%  
\\\hline
0.3 & 26.0\% & 65.4\% 
\\\hline
0.5 & 30.2\% & 68.5\% 
\\\hline
0.7 & 29.6\% & 68.6\% 
\\\hline
0.9 & 29.0\% & 68.6\%
\\\hline
\end{tabular}
}
\caption{Impact of Blending User Favorites Probability Distribution on the Accuracy Scores.}
\label{tab:offline_alpha_results}
\end{table}




\subsection{Inference Time and Size Analysis}


To measure the inference time of the models under test, we carry out an experiment where we run sentences via Android Jetpack Microbenchmark library and using the Samsung A23 with 8 CPU cores and 4GB RAM released in March 2022. We run these sentences with 300 warmup iteration and run these sentences for 50 times and take the average. Table \ref{tab:latency_results} shows that the median latency of all the models under test is within the acceptable 20 milliseconds range and sizes are insignificant as reported by \cite{hellsten2017transliterated, ramaswamy2019federated}.

 \begin{table}[http] \centering
\centering
{\small
\begin{tabular}{| c | c | c |}
\hline 
Model & Time (milliseconds) & Size (MB)
\\ \hline
90C & 19.3 & 2.8
\\\hline
590C & 20.4 & 3.1
\\\hline
1090C & 22.0 & 3.3
\\\hline
\end{tabular}
}
\caption{Inference Time and Size Comparison of Emoji Classifiers at Different Number Possible Output Classes.}
\label{tab:latency_results}
\end{table}






\subsection{Live Experiments}
\label{sec:liveResultsSec}

We ran a live-traffic experiment for users typing in English (US) only on SwiftKey Android cohort. We took a random sample of 75,000 users for each of the baseline and newly proposed 1090C model in this paper. The baseline relies on emoji to keyword associations and uses logistic regression. In the experiment, we used multiple success metrics to assess this experiment: precision, external emoji rate (ExRate), and click through rates (CTR). Precision is defined as the percentage of the emojis inserted from the top-1 emojis presented in the panel vs the total emojis inserted, while CTR looks at a user click from the top-24 emojis in the output panel. On the other hand, external emoji rate shows that the rate at which user inserts emojis from a third party application.

As shown in Table \ref{tab:live_results}, our 1090C model with augmentation leads to increased CTR, less external emoji rate, and improved precision of 12\%, 0.7\%, and 4\%, respectively relative to the logistic regression baseline. This performance is owing to the effectiveness of the 1090C model with augmentation and favorites reranking. Disregarding user preference personalization results in a 7\% drop in CTR recall, highlighting its importance.

\subsection{Alignment between Offline and Live Experiments}
\label{sec:liveResults2}

We observe that there is alignment between the offline results and live results to some great extent. In the live results case, we see that the increase in CTR amount is more than the precision, which is similar to the offline results shown in 
in Tables \ref{tab:offline_compare_hit1} and \ref{tab:offline_compare_hit24}. This is because CTR depends on the top-24 emoji predictions while precision depends on the top-1 emoji predictions, making it a stricter measure.




\begin{table}[ht]
\centering
\small
\begin{tabular}{|c|c|c|c|}
\hline
Model & Precision (\%) & ExRate & CTR \\
\hline
LogReg\_Baseline & 21.72\% & 3.83 & 7.3 \\
\hline
1090C\_Aug & 25.63\% & 3.80& 8.2 \\
\hline
\end{tabular}
\caption{Relative changes to metrics of the on-device classifier on live user traffic.}
\label{tab:live_results}
\end{table}

\begin{figure}[!ht]
  \includegraphics[width=0.5\textwidth]{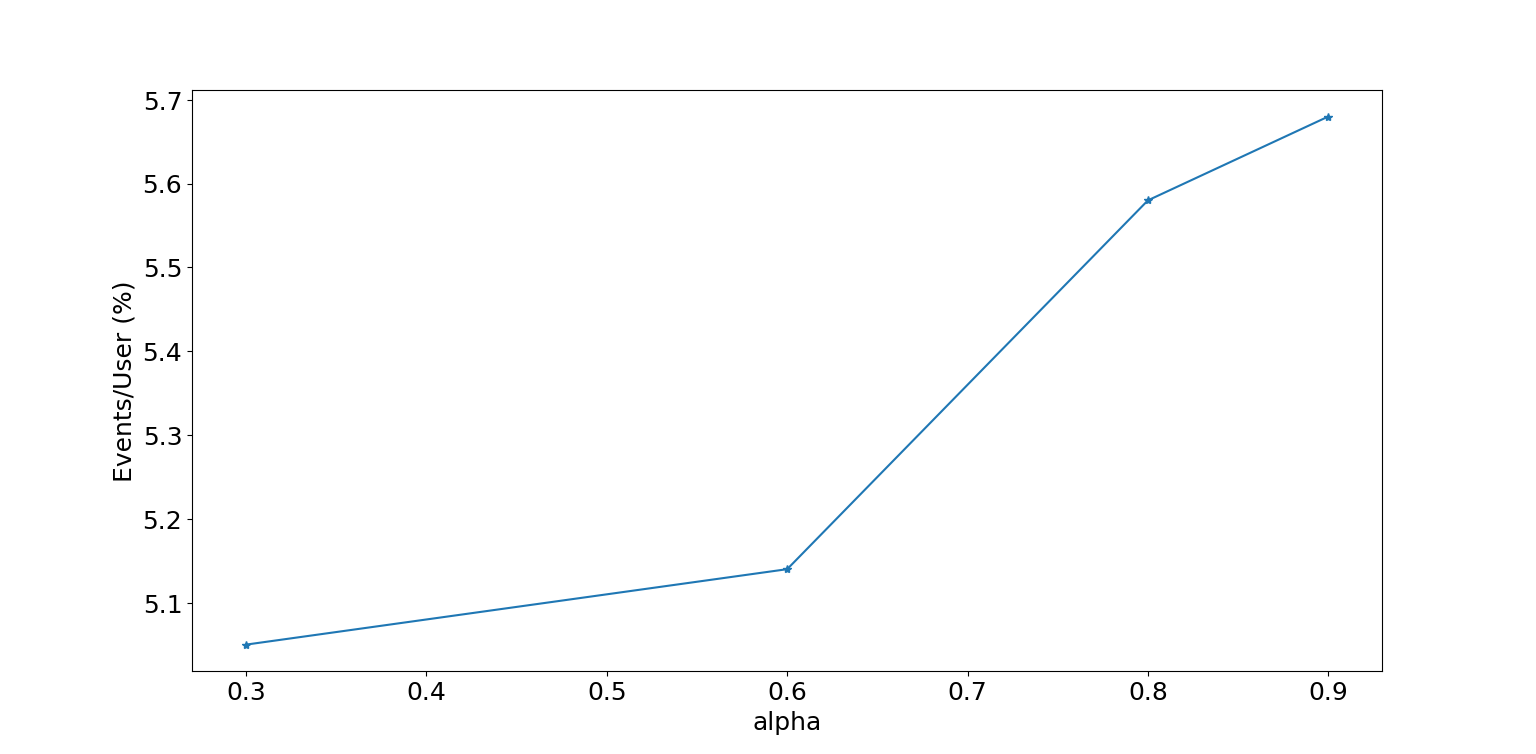}
  \caption{Impact of User Favorites reranked with Emoji Classifier on the Emoji Events per User on the Predictive Panel.}
  \label{fig:alpha_graph}
\end{figure}

As seen in the offline results with alpha in Table \ref{tab:offline_alpha_results}, live experiments also show that increasing alpha lead to increasing the total number of emoji insertion events per user as shown in Figure \ref{fig:alpha_graph}. This implies that the model probability distribution reranked slightly with a favorite probability lead to improved emoji predictive abilities. 






We carried out two user study experiments from SwiftKey users to compare the 90C with the 590C model, and 590C with the 1090C model. With 90C, we saw a CTR drop of 7\% because not all common emojis used by the user are included in this model. For 590C vs 1090C, results showed an improved CTR rate of 0.3\% because 1090C encompasses rarer emojis. These user study experiments justify the usage of 1090C model over the 90C model. All of these results together justify the deployment of the proposed on-device emoji classifer for SwiftKey.

\section{Conclusion}
\label{sec:conclusion}


This paper presents an on-device emoji classifier for SwiftKey trained with a new data augmentation method based on GPT that accounts for imbalanced emoji dataset distribution in numerous emoji classes. This classifier with GPT's data augmentation and method to adapt classifier output based on user preference increases the user engagement on SwiftKey's predictive panel and performs well particularly on rare emojis while keeping the model size and inference time intact to run on-device. These results show ways to alleviate the emoji imbalanced dataset distribution. Although emojis are commonly used in English according to internal benchmarks, we aim to extend this classifier to other languages and rich media.

\bibliography{custom}




\end{document}